\documentclass[]{spie}  
\usepackage{color}
\usepackage[]{graphicx}
\usepackage{amsfonts}

\title{Resolution enhancement in the recovery of underdrawings via style transfer by generative adversarial deep neural networks}

\author{George H.~Cann,\supit{a} Anthony Bourached,\supit{b} Ryan-Rhys Griffiths,\supit{c} and David G.~Stork\supit{d} 
\skiplinehalf 
\supit{a}Department of Space and Climate Physics, University College London, London, UK \\
\supit{b}Oxia Palus, London, UK \\
\supit{c}Department of Physics, University of Cambridge, Cambridge, UK \\
\supit{d}Portola Valley, CA 94028 USA}

\authorinfo{Send correspondence to the first author at {\tt george.cann.15@ucl.ac.uk}.}

\begin{document} 
\maketitle 

\begin{abstract}
We apply generative adversarial convolutional neural networks to the problem of style transfer to underdrawings and ghost-images in x-rays of fine art paintings with a special focus on enhancing their spatial resolution.  We build upon a neural architecture developed for the related problem of synthesizing high-resolution photo-realistic image from semantic label maps.  Our neural architecture achieves high resolution through a hierarchy of generators and discriminator sub-networks, working throughout a range of spatial resolutions.  This {\em coarse-to-fine} generator architecture can increase the effective resolution by a factor of eight in each spatial direction, or an overall increase in number of pixels by a factor of 64.  We also show that even just a few examples of human-generated image segmentations can greatly improve---qualitatively and quantitatively---the generated images.  We demonstrate our method on works such as Leonardo\rq s {\em Madonna of the carnation} and the underdrawing in his {\em Virgin of the rocks}, which pose several special problems in style transfer, including the paucity of representative works from which to learn and transfer style information.
\end{abstract}


\section{INTRODUCTION AND BACKGROUND} \label{sec:Intro}

Many paintings in the Western canon, particularly realist easel paintings from the Renaissance to the present, bear underdrawings and {\em pentimenti}---preliminary versions of the work created as the artist altered and developed into the final design.\cite{KammererZoldaSablatnig:03,Bomford:02,Andrzejewskietal:10}  In some cases the underdrawing represents a design separate from the final, visible work.  Such ghost-paintings appear in the oeuvre of artists such as Pablo Picasso, Vincent van Gogh, Rembrandt, and Francisco Goya, among many others.  Ghost paintings are most common in the early work of artists, when a financial strain may lead them to re-use canvases by painting over earlier designs.\cite{Kammereretal:04}  These ghost-paintings are revealed through x-radiography and infrared reflectography in conservation studios.\cite{Bomford:02}  

Scholars and the general art-loving public alike wish to view and study the hidden works as they were created---in full color and style---in order to get a richer understanding of the artist\rq s work and stylistic development.  The images revealed by such technical imaging present two difficult problems for subsequent scholarly analysis of the hidden work:

\begin{itemize}
\item These imaging methods mix or overlap the visible work with the ghost painting, so one must computationally separate the designs to isolate that of the hidden work.  A promising approach to this task relies on blind source separation, but is not the focus of our work presented here.\cite{ComonJutten:10}
\item These imaging methods produce grayscale (not color) images.  The richest images for scholarly analysis require the color---and more broadly speaking {\em style}---to be recovered.  This is the problem we address here:  computationally recovering, to the extent possible, the full color and style of the hidden artwork.  As part of that task, we address the sub-problem of enhancing the resolution of such a work, and we demonstrate that generative adversarial deep neural networks are effective in such ends.
\end{itemize}

In separate work, we have shown promising initial results for computational style transfer from representative artworks to such grayscale images of ghost-paintings.\cite{BourachedCann:19, BourachedCannGriffithsStork:21}  A drawback of that approach is that it generally leads to recovered images that are of low spatial resolution.  Often digital versions of x-rays or infra-red images of underdrawings are of spatial resolution too low for adequate scholarly analysis, however.  Our work presented below is centered on style transfer for grayscale edge maps, such as produced during the imaging of underdrawings, using a novel hierarchical deep network architecture to increase the final spatial resolution.  As an additional benefit our work here will enable comparisons between prior work using computational deep neural networks and generative adversarial neural networks for the problem of style transfer in ghost-painting recovery.\cite{Isolaetal:17}

In Sect.~\ref{sec:StyleTransfer} we briefly review methods for style transfer with particular attention to the application of recovering underdrawings and ghost paintings in fine art.  Then in Sect.~\ref{sec:StyleResults} we turn to our main concern:  neural techniques for enhancing the resolution of such images.  We present such enhanced images in Sect.~\ref{sec:StyleResults}, and summarize and outline several future directions in Sect.~\ref{sec:Summary}.

\section{Background:  Style transfer} \label{sec:StyleTransfer}

A number of contemporary methods for computationally recovering rich versions of such underdrawings based on style transfer using deep neural networks have been presented.\cite{GatysEckerBethge:16,Gatysetal:17,BourachedCannGriffithsStork:21}  While that parallel work demonstrated convincing style transfer, leading to computed images that are likely indicative of the original forms of the underdrawings, that research did not adequately address one component problem that will make the general technique more valuable to art scholars:  that is the problem of high spatial resolution.  
  
\begin{figure} 
\begin{center}
\includegraphics[width=.8\textwidth]{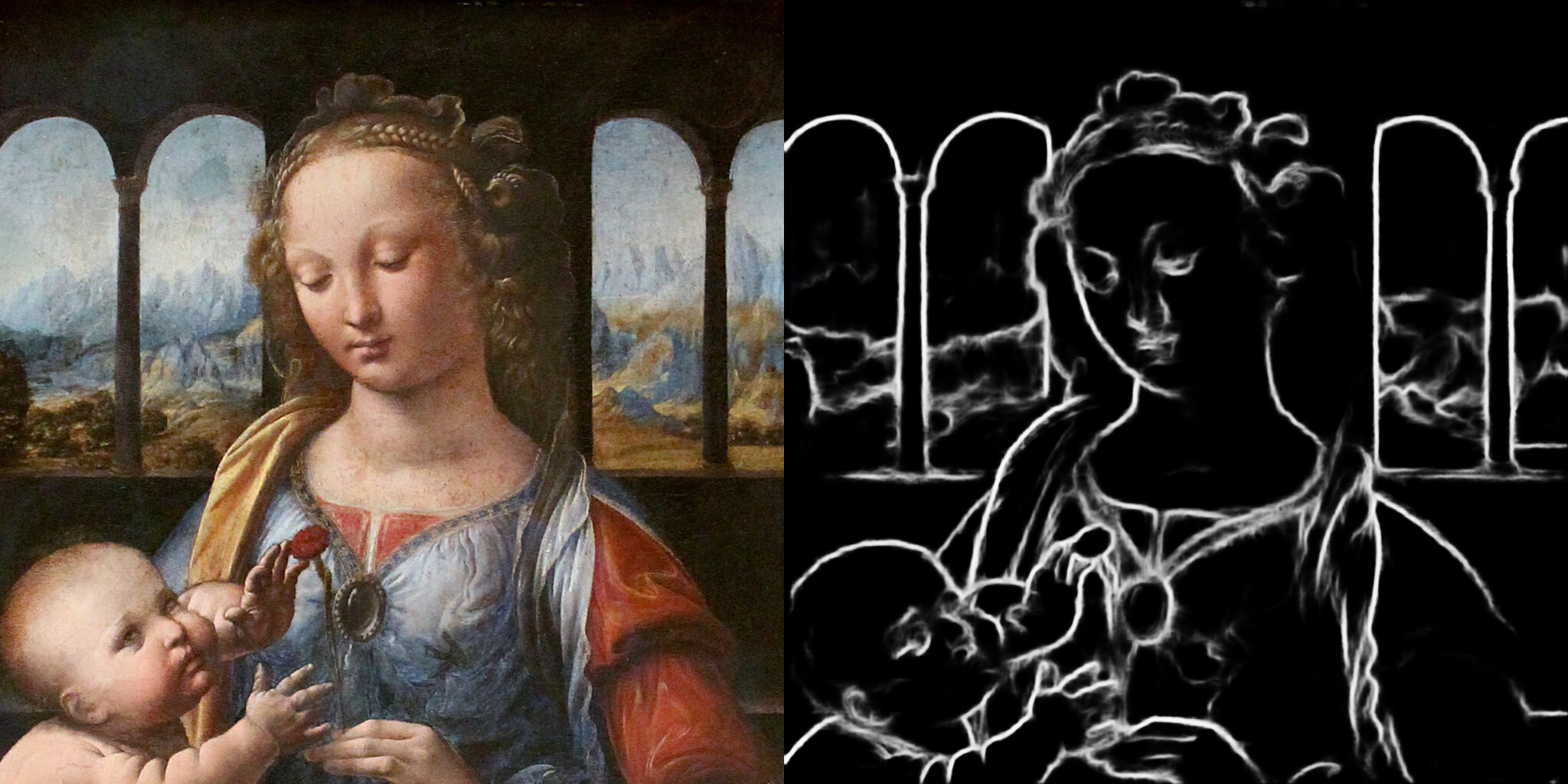}
\end{center}
\caption{({\sc L})~Leonardo\rq s {\em Madonna of the carnation} (or {\em Madonna with vase}), oil on wood panel (1478--80) (detail), and ({\sc R})~edge map, computed using deep network methods.  This edge map serves as a surrogate image of an underdrawing during the development and testing of our methods.\cite{XieTu:15}}
\label{fig:LeonardoMadonnaEdge}
\end{figure}

Specifically we approach the problem of computing recovered images having high resolution by means of a novel, powerful, hierarchical deep neural network architecture.  We formulate the computational problem as an adversarial minimax game of the form

\begin{equation} \label{eq:MinimaxGame}
\min\limits_G \max\limits_D {\cal L}_{cGAN} (G,D) ,
\end{equation}

\noindent between a {\em conditional generator network}, $G$, and a {\em discriminator network}, $D$.  (The subscript $cGAN$ indicates a conditional generative adversarial network.)\ \ We call the generator \lq\lq conditional\rq\rq\ because rather than mapping from a well-defined prior distribution of representative images (and hence styles), the network learns to generate conditional pairs of corresponding images, $\{ ({\bf s}_i , {\bf x}_i)\}$, from a training set.

The objective function in Eq.~\ref{eq:MinimaxGame} is:

\begin{equation} \label{eq:ObjectiveFunction}
{\cal L}_{cGAN} (G,D) = {\cal E}_{({\bf s}, {\bf x})} \left[ \log [D({\bf s}, {\bf x})] \right] + {\cal E}_{\bf s} \left[ \log [1 - D({\bf s}, G({\bf s}))] \right] ,
\end{equation}

\noindent where ${\cal E}$ is the expectation operator and its subscripts denote the domains of art images.  Together Eqs.~\ref{eq:MinimaxGame} and \ref{eq:ObjectiveFunction} represent a competition---or an adversarial \lq\lq game\rq\rq ---in which the conditional generator seeks to create an image similar to its set of representative images, while the discriminator seeks to enforce similarity to the incomplete underdrawing image.

\section{Resolution enhancement in style transfer} \label{sec:ResolutionEnhancement}

Our approach to spatial enhancement in the context of style transfer is to implement the minimax game throughout a hierarchy of paired generators and discriminators, which can be considered sub-networks.  Thus in the simplest case an overall generator network $G$ can be considered the serial composition of two sub-networks, that is, $G = \{ G_1, G_2 \}$.  (In principle we can employ more than two such stacked sub-networks.)\ \ Here $G_1$ works at a coarse scale (viz., up to size $1024 \times 512$ pixels), whose output is then passed to $G_2$, which acts as a local resolution enhancer, leading to a final high-resolution output (viz., $2048 \times 1024$ pixels).  Similarly, there are three serial discriminator sub-networks, $D = \{ D_1, D_2, D_3 \}$.  These discriminators are functionally the same but act at different downsampled scales:  $\times 1$, $\times 2$, and $\times 4$ by area.  Our overall architecture, then, encourages the coarse-to-fine objective of the generator, thereby enhancing the resolution of the final image beyond that of the original image.

\begin{figure} 
\begin{center}
\includegraphics[width=.8\textwidth]{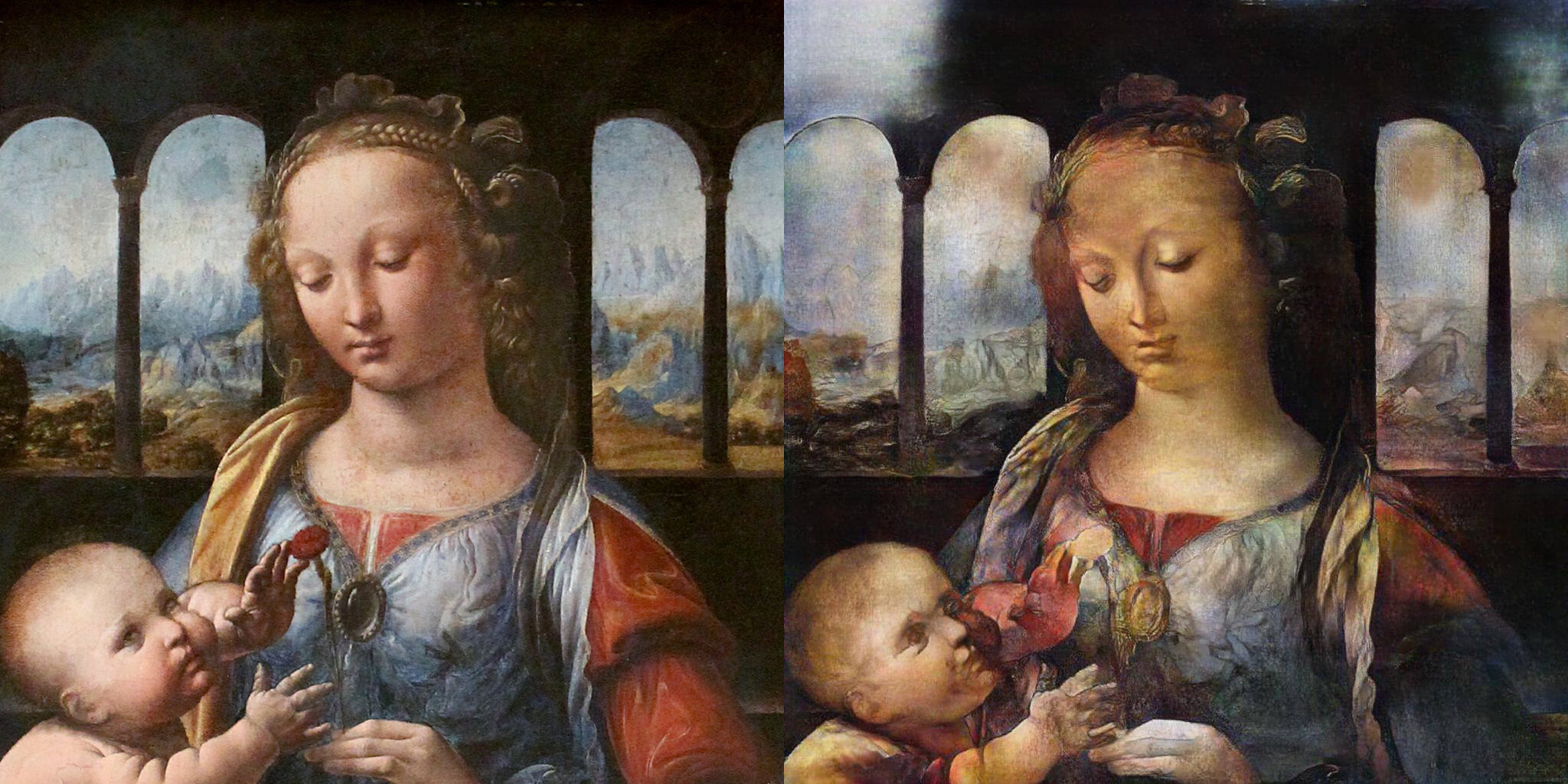}
\end{center}
\caption{({\sc L})~Leonardo\rq s {\em Madonna of the carnation} (or {\em Madonna with vase}), oil on wood panel (1478--80) (detail), and ({\sc R})~the \lq\lq recovered\rq\rq\ artwork generated from the edge map in the right panel of Fig.~\ref{fig:LeonardoMadonnaEdge} and a set of Lonardeschi paintings by means of our generalized adversarial neural network.  The peak signal-to-noise ratio, from Eq.~\ref{eq:PSNR}, is $PSNR \approx 15$, a high value indicating faithful recovery.}
\label{fig:LeonardoMadonnaInference}
\end{figure}

Our goal is to learn a mapping from a grayscale ghost-image x-ray fluorescence (XRF) or infrared reflectography (IRR) image of an underdrawing to a colored painting in oil.  Unlike the case in generating photorealistic image from semantically labeled maps, there is no ground truth pairing of $\{ ({\bf s}_i , {\bf x}_i) \}$.  That is, we have an underdrawing image, ${\bf s}_i$ from which we wish to infer the full, colored, source image ${\bf x}_i$, but we have no such ground-truth pairs.  (We can imagine {\em creating} a database of such pairs of images to form a ground-truth database, but such an effort is beyond the scope of our present work.)\ \ Nevertheless, we frequently do have a set of representative output images, $\{ {\bf x}_i \}$, associated with the artist in question.  Thus we can create a surrogate set of images, $\{ \tilde{\bf s}_i \}$, derived from $\{ {\bf x}_i \}$ in order to create a representative training set, $\{ (\tilde{\bf s}_i , {\bf x}_i) \}$.  The surrogate distribution, $\tilde{\bf s}_i$, of underdrawing images must represent as closely as possible the true distribution we seek to learn, ${\bf s}_i$.  

\subsection{Edge detection as preprocessing} \label{sec:EdgeDetection}

Our research approach is to use holistically-nested edge detection,\cite{XieTu:15} to create surrogate ghost-images as edge maps.  (Although we do not explore alternate methods for extracting such edge maps, we are confident that such alternatives would yield equally accurate, robust, and useful such maps.)\ \ We approximate the kinds of marks made by finite-size brushes by applying Gaussian pixel noise, $p \sim {\cal N}(0, 100)$ followed by a circularly-symmetric Gaussian blur kernel of width five pixels.  In this we we create a surrogate x-ray image of a ghost painting, and retain the (visible) colored image as ground truth for comparison and for quantifying the performance of our method.  

A great deal of information can be inferred from the edges of certain artworks.  For instance, artists such as Henri Matisse could depict full three-dimensional volumes and forms using sparse contour outlines, as revealed in his numerous line drawings, etchings, and paper cutout designs.  Full-color paintings contain shading and coloration information linked to such contour information and thus, in principle, bear visual information that can be transferred to images of just grayscale contours.  This is the information learned by our system.

Figure~\ref{fig:LeonardoMadonnaEdge} shows a detail from Leonardo\rq s {\em Madonna of the carnation} and the edge map produced by our method.  As mentioned, our central task is then to start with such an edge map, along with other, representative full-color paintings in the style of this artist, to compute an image that resembles the source painting or \lq\lq ground truth\rq\rq\ as closely as possible.  Our additional task is to increase the spatial resolution of the computed image above it corresponding source image.

There remains two methodological challenges to our approach:

\begin{itemize}
\item The dataset of available \lq\lq style\rq\rq\ images is often small.  In particular case, Leonardo completed only about three dozen easel paintings.
\item Our inference will be based on a distribution of underdrawings that differs from surviving paintings, that is, the priors over densities obey $p({\bf s}) \neq p(\tilde{\bf s})$, in general.
\end{itemize}

We addressed the first problem by using paintings from the broader set of Leonardeschi paintings, that is, paintings from the followers of Leonardo, such as Giovanni Antonio Boltraffio, Ambrogio de Predis, Francesco Napoletano, Andrea Solario, and Giacomo Caprotti, the last artist is known more broadly as Sala{\'i}, Leonardo\rq s younger, impish long-time confidant.  

\begin{figure} 
\begin{center}
\includegraphics[width=.8\textwidth]{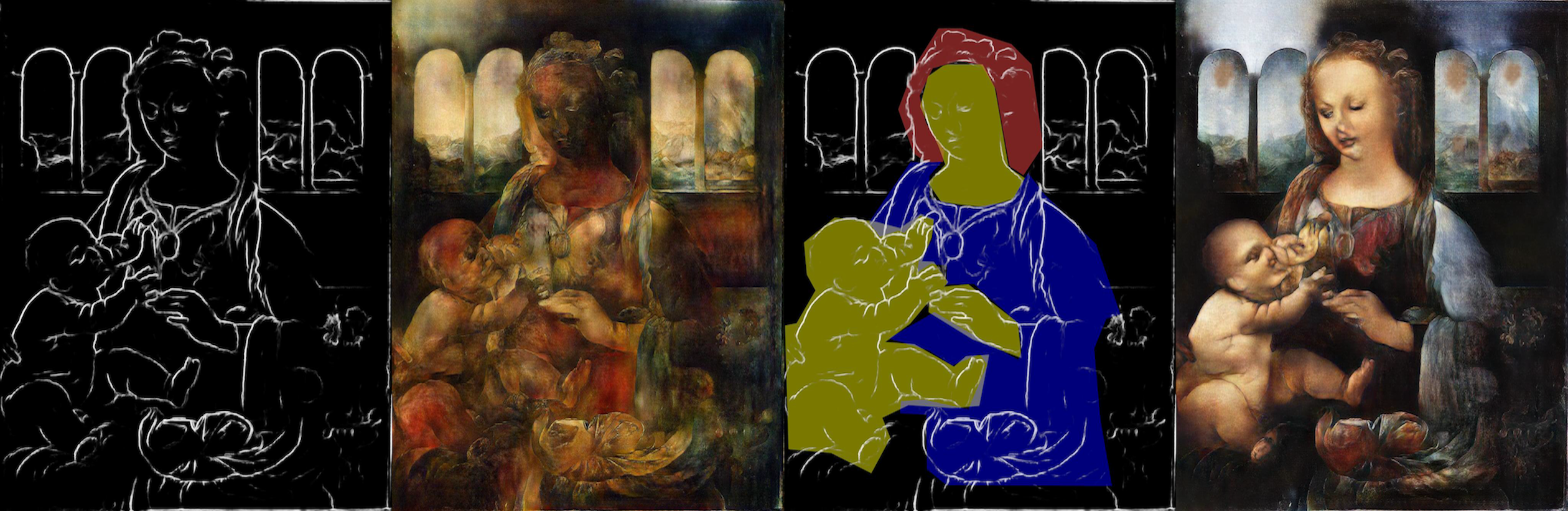} \\
a) \hspace{2.6cm} b) \hspace{2.6cm} c) \hspace{2.6cm} d)  
\end{center}
\caption{a)~The computed edge map for Leonardo\rq s {\em Madonna of the carnation}, reproduced from Fig.~\ref{fig:LeonardoMadonnaEdge} to aid visual comparisons here.  b)~The preliminary results of style transfer using the convolutional generalized adversarial network.  Although the overall color and lightness schemes approximate well the ground truth painting in Fig.~\ref{fig:LeonardoMadonnaEdge}, the rendering of much of the Virgin\rq s costume and skin tones throughout are severely mottled and do not match the ground truth.  c)~Human-generated coarse segmentation of skin and hair regions.  Notice that the entered regions consist of coarse polygons and as such do not conform closely to the natural contours in the work.  d)~The recovered artwork inferred from the style images as well as the single human-generated segmentation information in c).  Here the skin tones closely resemble the ground truth data, as does the image overall.}
\label{fig:LeonardoMadonnaLabelNonlabel}
\end{figure}

\subsection{Low-to-high resolution} \label{sec:low-to-high}

As mentioned above, a key problem is increasing the resolution of a recovered image, preferably to a level of use to art scholars.\cite{KimLeeLee:16}  We addressed this problem by starting with a high-resolution image of Leonardo\rq s {\em Mona Lisa}, of dimensions $8192 \times 12288$ pixels, from which we created $96$ non-overlapping tiles, each of which was $1024 \times 1024$ pixels.  Each such patch was digitally filtered so as to eliminate the highest-resolution noise, such as craquelure.  These new images, then, formed a new set of conditional pairs of images, $\{ ({\bf s}_i , {\bf x}_i )\}$.  We trained a superresolution model of two cascading sub-networks with such paired images, thereby learning a map from a lower-resolution image to a higher-resolution image, as governed by Eqs.~\ref{eq:MinimaxGame} and \ref{eq:ObjectiveFunction}.  The resulting high-resolution image does not include the noise of cracks and such.

\begin{figure} 
\begin{center}
\includegraphics[width=.8\textwidth]{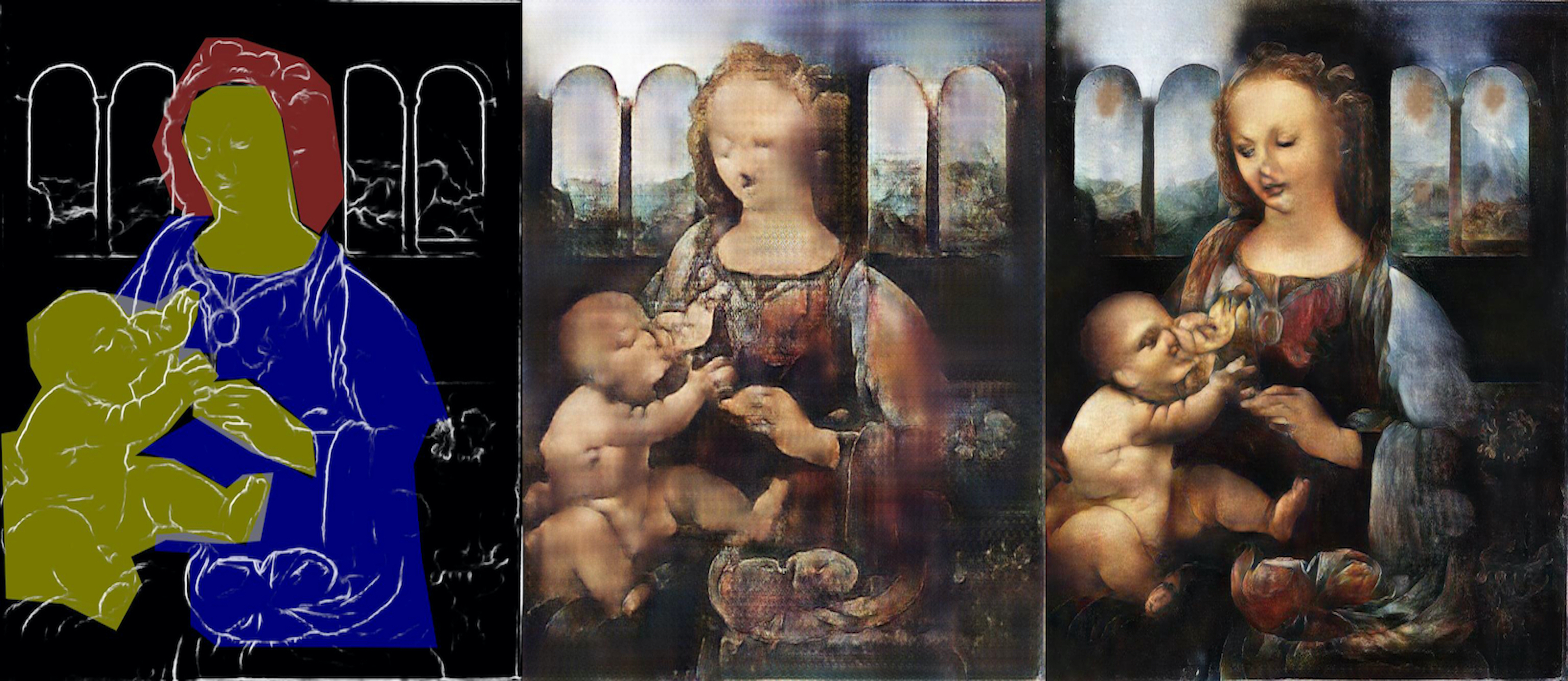} \\
a) \hspace{1.7in} b) \hspace{1.7in} c)
\end{center}
\caption{a)~The coarse human-generated image segmentation of a detail of Leonardo\rq s {\em Virgin of the carnation}.  b)~The computed image reconstruction without the segmentation information from a), and c)~ the computed reconstruction using the segmentation in formation from a).  Clearly this segmentation information leads to a more faithful and accurate reconstruction, particularly in the skin of both figures.}
\label{fig:SameInputBetterTexture}
\end{figure}

The final and complete image was computed by upsampling the $1024 \times 1024$-pixel patches to $8192 \times 8192$ pixels---a factor of eight in each linear dimension.  Our initial efforts were based on non-overlapping image patches, and these led to slight discontinuities and block artifacts.  One approach to reducing such image artifacts would be to include overlapping boundaries between component blocks.  We explored an alternate approach in order to avoid such artifacts:  we overlapped segments {\em throughout} image patches.  Specifically, we split blocks at every $64$ pixels, thereby creating $\left( \frac{8192 - 1024}{64} \right)^2 = 12544$ overlapping segments.

\begin{figure} 
\begin{center}
\includegraphics[width=0.95\textwidth]{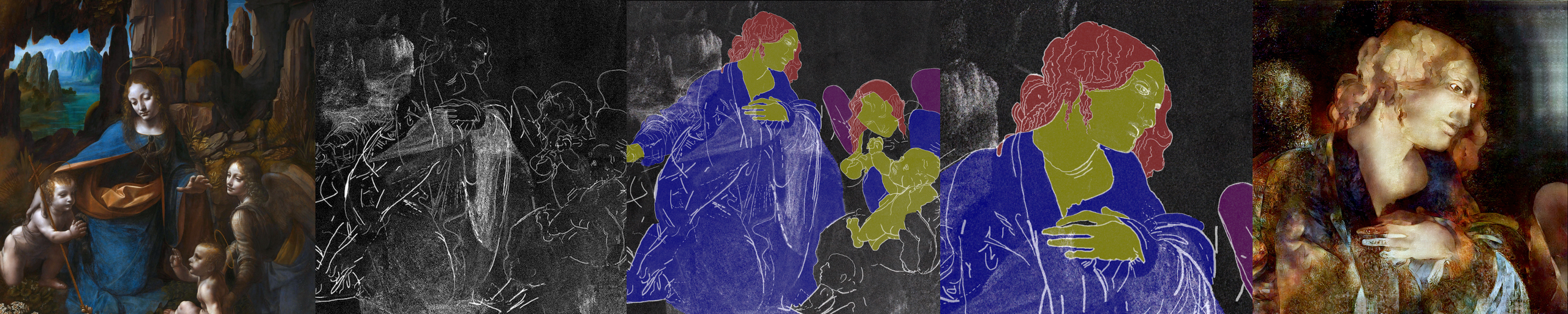} \\
a) \hspace{2.7cm} b) \hspace{2.7cm} d) \hspace{2.7cm} d) \hspace{2.7cm} e) 
\end{center}
\caption{a)~Leonardo\rq s {\em Virgin of the rocks} ($189.5 \times 120\ cm$), oil on panel (1495--1508), National Gallery London.  b)~The grayscale underdrawing of a Madonna and child composition, revealed by x-radiography.  c)~Human-generated coarse segmentation of the underdrawing.  d)~A detail showing just the Virgin\rq s head from panel c).  e)~The computed full-color ghost-painting of the passage in d).}
\label{fig:LeonardoAngelRocksProgression}
\end{figure}

\begin{figure}
\begin{center}
\includegraphics[width=0.9\textwidth]{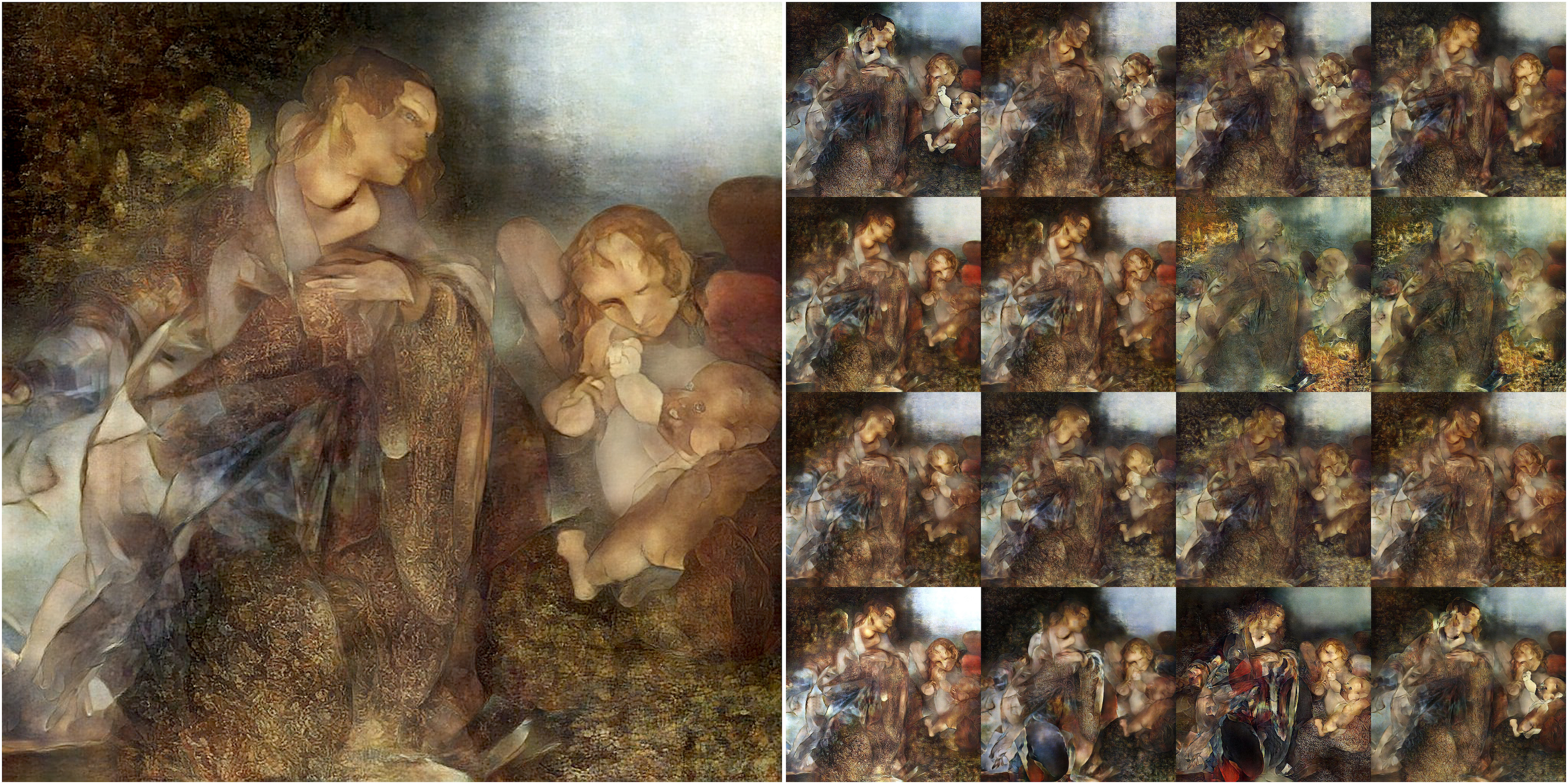} \\
a) \hspace{7cm} b) \hspace{7cm} 
\end{center}
\caption{a) The full inferred ghost-painting behind Leonardo\rq s {\em Virgin of the rocks}, computed by our convolutional generative adversarial networks.  Clearly the style does not fully represent that of Leonardo:  the contours are rather sinuous and the skin modeling is uneven and mottled, unlike the sfumato and chiaroscuro that characterizes his portraits. b) A set of 16 inferred works used to derive a).}
\label{fig:LeonardoAngelRocksBehind}
\end{figure}

\begin{figure} 
\begin{center}
\includegraphics[width=.95\textwidth]{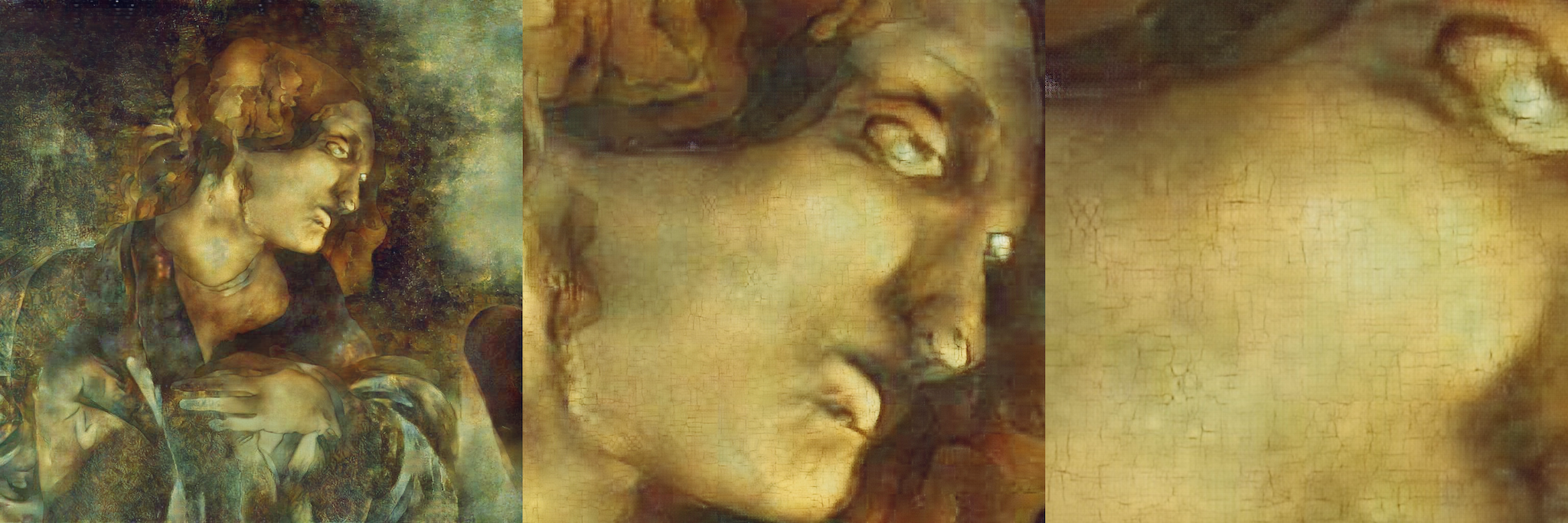}
\end{center}
\caption{An example superresolution of Fig.~\ref{fig:LeonardoAngelRocksProgression} d) showing the reconstructed face.}
\label{fig:LeonardoZoomedMultires}
\end{figure}

We then performed superresolution inference on each segment and numerically averaged along the appropriate lattice points in order to compute the final image.  Such averaging reduces the previous artifacts significantly.  Fig.~\ref{fig:LeonardoAngelRocksBehind} shows before superresolution inference, and Fig.~\ref{fig:LeonardoZoomedMultires} shows after such superresolution inference.  The color style is, qualitatively speaking, more similar to the {\em Mona Lisa} on which the superresolution model was trained.

\section{Spatial enhancement in style transfer:  Results} \label{sec:StyleResults}

Figure~\ref{fig:LeonardoMadonnaInference} shows representative preliminary results for Leonardo\rq s double portrait.  The shading and coloration is generally coherent and consistent though the skin passages are somewhat mottled and lack the sfumato and chiaroscuro for which this artist is celebrated.  Notice, though, that the colors of the Virgin\rq s costume match well, though the yellow cloth over her shoulder at the left.

We hypothesized that the mottling of skin tones in Fig.~\ref{fig:LeonardoMadonnaInference} was due to inaccuracies in segmentation---specifically that the database of Leonardeschi \lq\lq style\rq\rq\ images represented skin regions did not conform accurately to the contours in the original painting.  The most principled solution to this problem would be to increase the number of representative images for training.  Unfortunately, we could not take this approach because the number of such paintings relevant to this case is rather low.  Our alternative was to exploit human expertise and knowledge, specifically about segmentation by means of {\em semi-semantic semi-supervised labelling} (SSSSL) based on human labeling of regions.  We marked regions by hand, using low opacity, based on their broad segmentation categories of skin, hair, clothes, or wings without fine internal detail, as shown in the third panel of Fig.~\ref{fig:LeonardoMadonnaLabelNonlabel}.  

The third panel in Fig.~\ref{fig:LeonardoMadonnaLabelNonlabel} shows the improvement in rendered images afforded by such human-generated coarse segmentation information.  The difference between inference using such SSSSL and unlabeled images in the regime of small data sets.  These conditions most naturally represent the out-of-distribution performance.  Note that the clear superiority of generalization using SSSSL in that figure, indicating that a small amount of information included by hand leads to a marked improvement in rendered images.  

We can quantify the benefit of such coarse segmentation information using the peak signal-to-noise figure of merit---a measure of the similarity of a computed image to the ground truth image.  To this end, we first define the pixel-wise mean-squared error as:

\begin{equation} \label{eq:MSE}
{\rm MSE} = \frac{1}{3 N M} \sum\limits_{i=1}^N \sum\limits_{j=1}^M \left( (r_{ij} - \hat{r}_{ij})^2 + (g_{ij} - \hat{g}_{ij})^2 + (b_{ij} - \hat{b}_{ij})^2 \right) ,
\end{equation}

\noindent where in our case each color channel ($r$, $g$, $b$) is represented with eight bits, corresponding to a range of $0 \to 255$.  The peak signal-to-noise ratio is then:

\begin{equation} \label{eq:PSNR}
{\rm PSNR} = 20 \cdot \log_{10} \left( \frac{255}{\sqrt{\rm MSE}} \right) .
\end{equation}

\noindent This metric measures, in bits, a signal-to-noise ratio, so the higher its value the more faithful the resulting image.

\begin{figure} 
\begin{center}
\includegraphics[width=.7\textwidth]{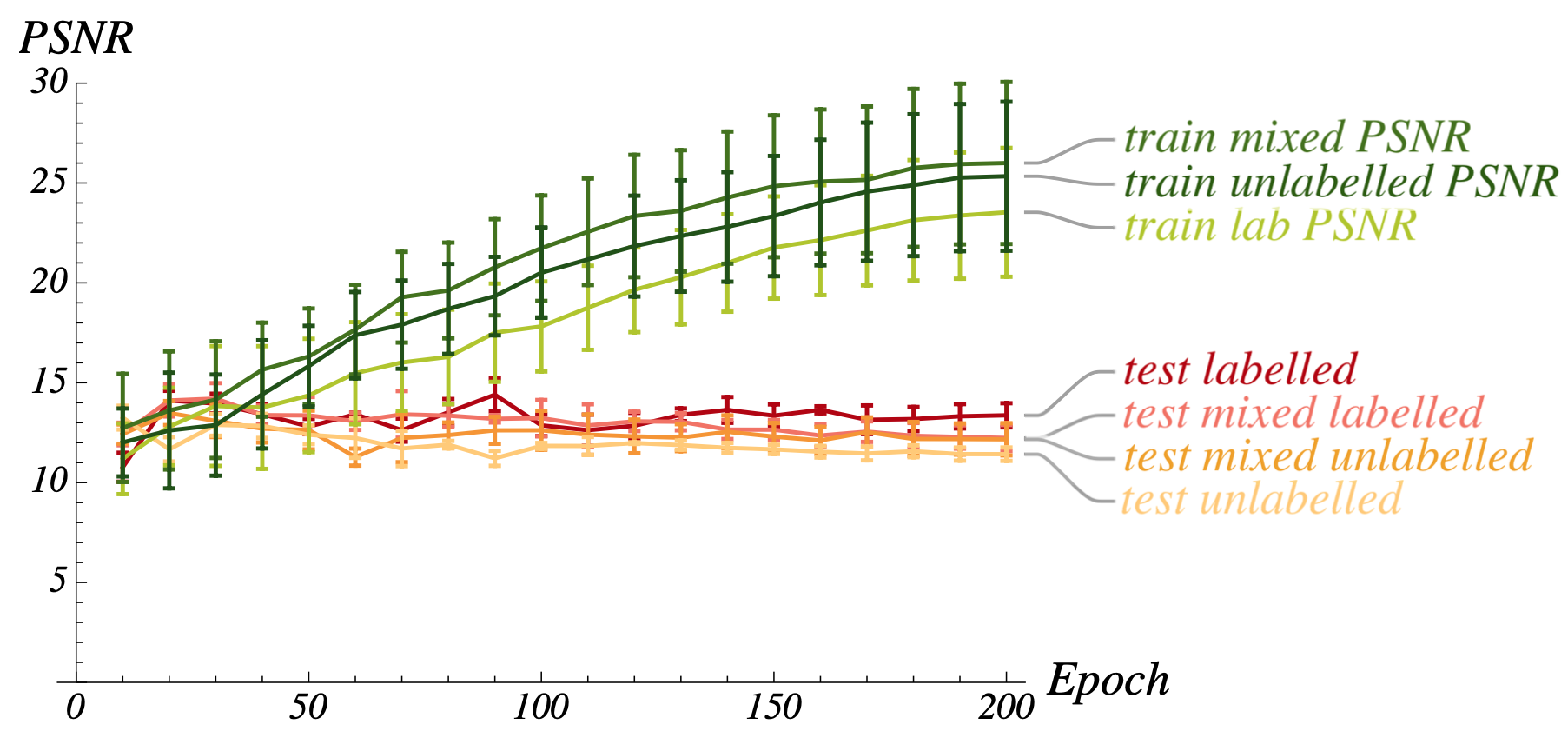}
\end{center}
\caption{The peak signal-to-noise ratio, as defined in Eq.~\ref{eq:PSNR}, for image data of various forms.  The data including human-generated (coarse) segmentation information (top curve) is consistently and statistically significantly higher (\lq\lq better\rq\rq ) than all other cases.}
\label{fig:PSNRTestTrain}
\end{figure}

Such improvement can be quantified using the peak signal-to-noise ratio, or PSNR of Eq.~\ref{eq:PSNR}.  Figure~\ref{fig:PSNRTestTrain} shows the PSNR during training of the test images in Fig.~\ref{fig:LeonardoMadonnaLabelNonlabel}.  The performance on the test set is consistently higher when using SSSSL and the learning curve plateaus quickly, just as we would expect.  After all, the segmentation information provided by the human experts---even though rather coarse---nevertheless constrains the computed image, leading to a more accurate and faithful image.

The x-ray revealing the underdrawing in Leonardo\rq s {\em Virgin of the rocks} presents a special challenge to our approach.  The contour of the underdrawing is rather complex, with numerous regions with details and oblique angles.  Figure~\ref{fig:LeonardoAngelRocksProgression} shows the progression, including a detail of the Virgin\rq s head in the right panel.

\section{Summary} \label{sec:Summary}

We have demonstrated that grayscale edge maps of underdrawings, such as provided by x-radiography and infrared reflectography, can be colored through style mapping of appropriate paintings by means of genererative adversarial neural networks.  The resolution of the final images can be enhanced through our novel stacked sub-network architecture, each sub-network leads to an an effective spatial oversampling.  We have shown, moreover, that coarse human-generated segmentation information---specifically related to skin regions---can lead to dramatic improvements in the accuracy of the style of computed images.

\begin{figure} 
\begin{center}
\includegraphics[width=.9\textwidth]{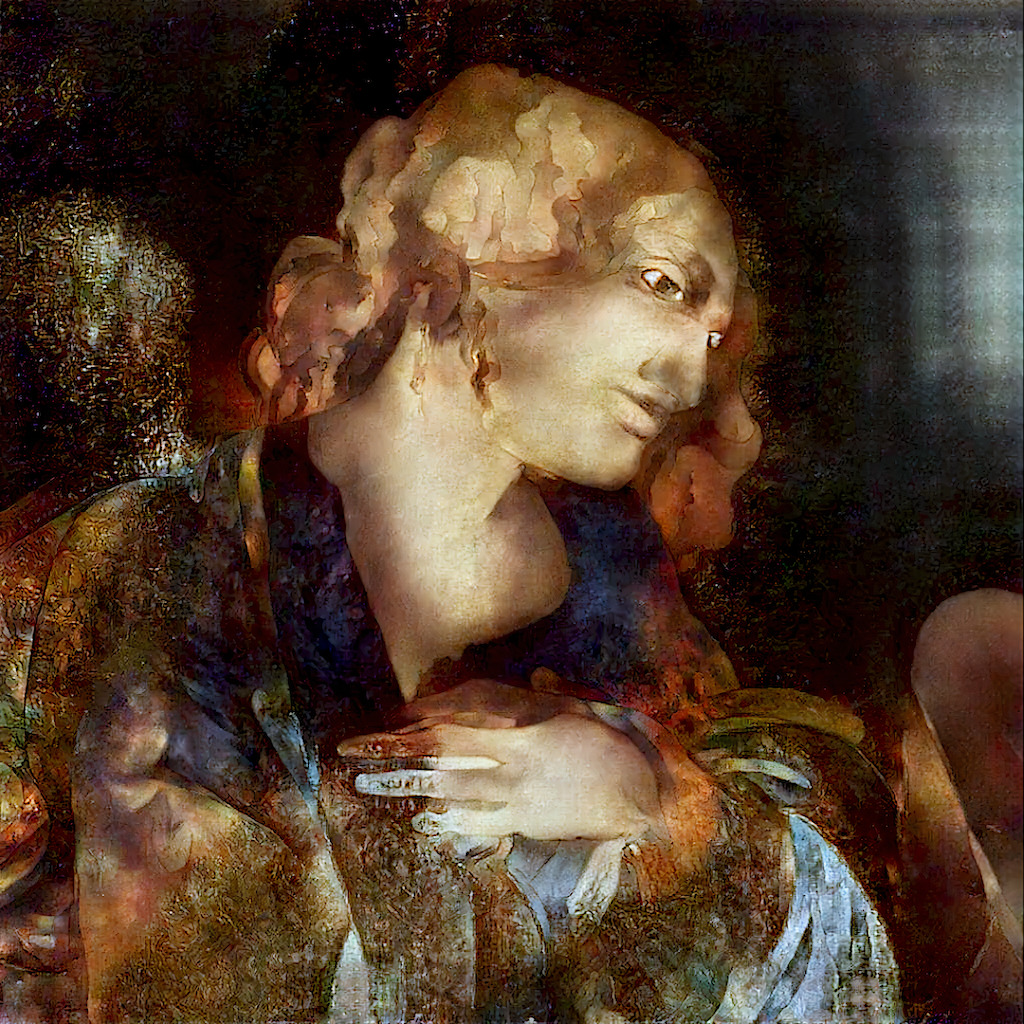}
\end{center}
\caption{The reconstructed face of Madonna.}
\label{fig:LeonardoReconstructedFace}
\end{figure}

Although this work has validated our general approach, our results are not yet sufficiently accurate for careful art-historical analysis.  We see several general avenues of research needed to achieve such ends.  First, we need larger corpora of representative artworks for style transfer.  The size of such databases is determined primarily by the accidents of art history:  how many artists created appropriate works, how many such works were executed, and how many survive.\cite{Charney:18}  It may be that works from some artists or periods simply lack adequately large such corpora and computational methods will be of negligible value.  Second, we may incorporate large databases of human-generated constraint information.  Here natural human-machine interfaces will be an asset so that art experts or even the general public can easily provide this information.  We can imagine broad knowledge capture systems over the web, such as the Open Mind Initiative, Mechanical Turk, and their many descendants.  Third, we envision improvements to the network architectures and learning protocols, specifically ones tailored to the problems of learning and inferring two-dimensional region texture and color from one-dimensional contours.  Of course the final stage is to package these digital tools in a form for art scholars, particularly those who may not be also computer scientists.\cite{Wangetal:18,Reinhardetal:01,Lietal:17,Selimetal:16,Zhangetal:13,TenenbaumFreeman:00}

It appears that such improvements on the work presented here will lead to digital tools that will find use by art scholars, particularly conservators, curators, and technical image analysts.

\section*{ACKNOWLEDGEMENTS}

We would like to thank the National Gallery London, home of Leonardo\rq s {\em Virgin of the rocks}, and the Alte Pinokothek, Munich, home of Leonardo\rq s {\em Virgin of the carnation}.  The last author would like to thank the Getty Research Center for access to its Research Library, where some of the above research was conducted.

\bibliography{Art}
\bibliographystyle{spiebib}
 
\end{document}